\documentclass[11pt]{article}

\usepackage[final]{acl}

\usepackage{times}
\usepackage{latexsym}

\usepackage[T1]{fontenc}

\usepackage[utf8]{inputenc}

\usepackage{microtype}

\usepackage{inconsolata}

\usepackage{graphicx}

\usepackage{booktabs}
\usepackage{multirow} 
\usepackage{amsmath}
\usepackage{amssymb}

\usepackage[table]{xcolor}
\usepackage{array}
\usepackage{enumitem}
\usepackage{pifont}      
\usepackage{fontawesome5}

\newcommand{\model}{\texttt{{EGCForest}}}

\newcommand{\datasettitle}[1]{%
  \par\noindent
  {\normalsize\bfseries #1\par}%
  \vspace{0.3ex}
}

\newcommand{\methodexample}[2]{%
  \par\smallskip\noindent
  \begingroup
  \setlength{\fboxsep}{5pt}%
  \colorbox{black!5}{%
    \begin{minipage}{\dimexpr\linewidth-2\fboxsep\relax}
      \small\setlength{\parindent}{0pt}%
      \textbf{#1}\par\smallskip
      #2
    \end{minipage}%
  }%
  \endgroup
  \par\smallskip
}

\definecolor{timeblue}{RGB}{230,242,255}
\definecolor{tokengreen}{RGB}{232,248,239}
\definecolor{storagegray}{RGB}{245,245,245}

\title{EGCForest: Structured Agent Memory Organization for Enterprise Group Chats}

\author{
  \textbf{Lingxi Cui\textsuperscript{1,2}},
  \textbf{Jinhao Ruan\textsuperscript{1,2}},  
  \textbf{Wenqing Wei\textsuperscript{1,2}}, 
  \textbf{Guo Yi\textsuperscript{3}}, 
  \textbf{Li Rao\textsuperscript{3}},
  \textbf{Ke Chen\textsuperscript{3},}
  \textbf{Lidan Shou\textsuperscript{1,2\ding{41}}}
  \\
  \textsuperscript{1}The State Key Laboratory of Blockchain and Data Security, Zhejiang University \\
  \textsuperscript{2}Hangzhou High-Tech Zone (Binjiang) Institute of Blockchain and Data Security \\
  \textsuperscript{3}Xiaohongshu \\
  \normalsize{\tt\{cuilingxi.cs, capt, chenk, should\}@zju.edu.cn}\\
  \normalsize{\tt\{zengzicheng, yutian5, liuxin6\}@xiaohongshu.com}
}

\begin{document}
\maketitle

\begin{abstract}
Enterprise group chats record project progress, decisions, task ownership, and other organizational information valuable to enterprise agents.
Reliable retrieval is challenging because evidence spans interleaved threads, channels, and time, with meaning and authority conditioned on the speaker.
Neither RAG's independent chunk retrieval nor memory systems' summary-based consolidation can reliably recover such distributed evidence.
We present \model{}, a novel multi-view hierarchical memory framework for enterprise group chats.
\model{} organizes an immutable message log into three complementary forests:
a Channel forest for temporal workflows, an Actor forest for speaker-conditioned context, and an Artifact forest for deliverable lifecycles.
Cross-view connections are enabled by a Placement Map and sparse structural bridges. 
Built on this memory backbone, \model{} retrieves and assembles evidence by jointly leveraging message-level relevance, hierarchical context from forest summaries, and cross-view links, recovering distributed evidence.
Experiments on three enterprise-chat benchmarks show
that \model{} consistently outperforms RAG and memory-system baselines, improving answer accuracy by up to 20.85 points over MemOS while supporting efficient online retrieval.
The code is available at \url{https://anonymous.4open.science/r/EGCForest-MAD12/}.
\end{abstract}

\section{Introduction}
\label{sec:intro}

Enterprise group chats have become the default medium for daily collaboration inside modern organizations~\cite{yang2026groupmembench}. Beyond casual exchanges, they carry a substantial amount of task-related knowledge such as project progress, ownership assignments, and design rationales. 
Recent progress in agent systems has raised the demand for reliable enterprise knowledge sources~\cite{sun2026EnterpriseKnowledge}, and group-chat logs provide large-scale, fine-grained evidence of organizational activity.
Existing research on group chats, in contrast, has largely focused on \emph{participating} in the conversation~\cite{sapkota2025multi}, for instance predicting the next speaker, recognizing the addressee, or disentangling interleaved threads, rather than on \emph{extracting} the organizational knowledge that these conversations contain. Turning raw chat logs into an external memory that an agent can query accurately, with traceable evidence back to the original messages, therefore remains an open problem.

\begin{figure}[!t]
\centering
\includegraphics[width=\linewidth]{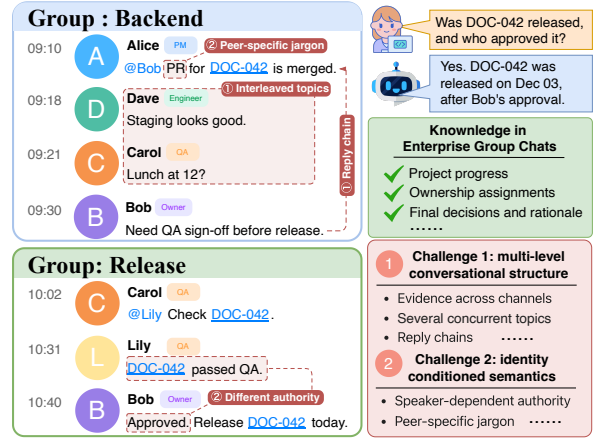}
\caption{Enterprise group chats contain valuable organizational knowledge, but deploying them as reliable agent memory is challenging.}
\label{fig:motivating}
\end{figure}

Two properties of enterprise group chats make this task particularly challenging. 
First, reliable memory for enterprise group chats must preserve their \textbf{multi-level conversational structure}.
Messages are temporally ordered, but topics interleave within channels, reply chains mix substantive and tangential exchanges, and decision evidence may span multiple threads, channels, and time periods.
As shown in Figure~\ref{fig:motivating}, answering an artifact-status question requires tracing its lifecycle across two channels and four participants.
Retrieval-augmented generation (RAG) systems typically treat messages as independent chunks, favoring locally similar matches while missing the full evidence chain~\cite{karpukhin2020dense}.
Memory systems instead compress chats into facts or summaries, obscuring or conflating evidence from distinct threads and decisions~\cite{chhikara2025mem0}.

Second, capturing \textbf{speaker-conditioned semantic}s is challenging yet essential in enterprise group chats. 
Unlike dyadic conversations, single-user agent interactions, or participant-agnostic documents, group-chat participants' roles, responsibilities, and task relations shape message interpretation.
The same claim may carry different authority across speakers (e.g., a project owner versus an intern), while teams may refer to the same artifact differently.
RAG systems typically represent messages as plain text and use speaker identity only as a metadata filter, thereby overlooking speaker-conditioned meaning and evidential relevance~\cite{maharana2024evaluating}.
Memory systems often encode identity as a namespace or auxiliary tag, which can similarly conflate claims from participants with different responsibilities~\cite{packer2023memgpt}.

We address both challenges with \model{}, a structured external memory organized as three complementary forests over a shared message log. 
The \textbf{Channel} forest partitions each group chat into hierarchical temporal intervals, preserving local context and topic evolution.
The \textbf{Actor} forest centers each participant, with separate branches for messages authored by that participant and messages that mention them, making identity a first-class semantic axis rather than a passive metadata field. 
The \textbf{Artifact} forest organizes messages around deliverables and represents their lifecycle across group chats.
These forests are explicitly interconnected. When a message appears in multiple forests, a \textbf{Placement Map} links its view-specific placements to a single immutable log entry, enabling direct cross-view correspondence without content duplication.
\model{} further adds sparse \textbf{Bridges} that connect distinct messages through native reply links and shared salient identifiers.
Built on this backbone, a four-stage retrieval pipeline combines view-aware leaf recall, summary-guided supplementation, structural completion, and budget-aware assembly.

\smallskip
\noindent \textbf{Contributions.} We make the following contributions:
\textbf{(1) A novel memory framework.} We propose \model{}, a structured memory framework featuring three complementary forests to organize scattered knowledge from enterprise group chats while preserving conversational structure, speaker context, and traceability to original messages.
\textbf{(2) Three-view forest architecture.} We construct complementary Channel, Actor, and Artifact forests that capture temporal workflow, speaker-conditioned context, and artifact lifecycles, respectively, with a Placement Map and sparse Bridges enabling efficient cross-view navigation.
\textbf{(3) Structure-aware evidence retrieval.} We combine view-aware tree routing and evidence expansion through summaries and cross-view links to recover distributed evidence.
\textbf{(4) Empirical validation.} Across three benchmarks, \model{} delivers consistent improvements over strong RAG and memory-system baselines, outperforming \texttt{MemOS} by up to 20.85 points.

\section{Preliminaries}
\label{sec:preliminary}

\subsection{Problem Formulation}
\label{sec:problem}

We consider an enterprise workspace with a set of channels\footnote{We use \emph{channel} to refer to a persistent, multi-party group chat, following common enterprise chat platform terminology (e.g., Slack, Microsoft Teams).}
$\mathcal{C}$ and a set of users $\mathcal{U}$. 
A chat history is an ordered, immutable sequence of messages $\mathcal{M} = \{m_1, m_2, \ldots\}$, where each message is a tuple $m = (t_m, u_m, c_m, r_m, x_m)$ with timestamp $t_m$, sender $u_m \in \mathcal{U}$, channel $c_m \in \mathcal{C}$, optional reply target $r_m \in \mathcal{M} \cup \{\emptyset\}$, and raw text $x_m$.

Our goal is to organize $\mathcal{M}$ into a structured external memory
$\mathcal{K}$ that an agent can query. As memory quality is hard to assess
directly, we use question answering as a measurable proxy for it. 
Concretely,
given a natural-language query $q$ posed by an asking user $u_q \in \mathcal{U}$,
the task is to retrieve a compact evidence set $\mathcal{E}_q = \mathcal{R}(q, u_q; \mathcal{K}) \subseteq \mathcal{M}$
via memory $\mathcal{K}$, 
and feed $\mathcal{E}_q$ together with $q$ to an
answer agent to produce an answer $a$.

\begin{figure*}[t]
\centering
\includegraphics[width=\linewidth]{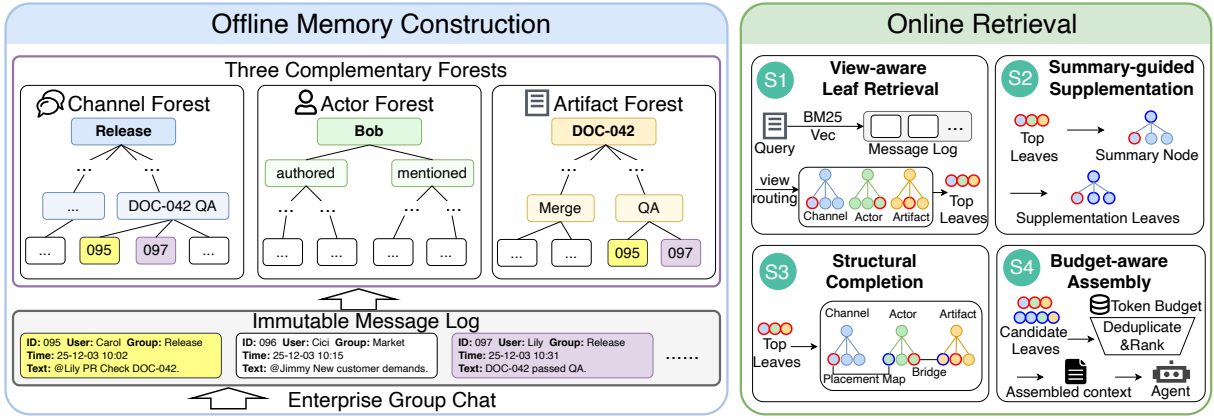}
\caption{
Overview of \model{}: offline three-view memory construction and online four-stage retrieval.
}
\label{fig:framework}
\end{figure*}

\subsection{Related Work}
\label{sec:related}

\textbf{Agent memory systems} maintain persistent memory across sessions
with a single user, typically by compressing conversations into facts,
summaries, or memory items~\cite{xu2026mem,kang2025memoryos,latimer2025hindsight}.
MemGPT~\cite{packer2023memgpt} treats the context window as limited main
memory and swaps information with an external store, whereas
Mem0~\cite{chhikara2025mem0} incrementally adds, merges, or replaces memory
entries. Although effective for single-user assistants, these
compression-oriented designs can obscure group-chat structure by treating
speaker identity as metadata, merging conflicting
claims, and limiting access to raw-message evidence.

\textbf{Retrieval-augmented generation (RAG)} typically organizes text corpora as flat text chunks, hierarchical summaries, or graph-structured representations.
BM25~\citep{robertson2009probabilistic} and dense retrieval~\citep{karpukhin2020dense} rank individual chunks by lexical or semantic relevance.
Hierarchical methods such as RAPTOR~\cite{sarthi2024raptor} recursively cluster and summarize chunks or retrieval across abstraction levels. 
GraphRAG~\citep{edge2024graphrag}, LightRAG~\citep{guo2025lightrag}, and
HippoRAG~\citep{gutierrez2024hipporag} organize entities and relations for retrieval, with GraphRAG additionally summarizing graph communities. 
These approaches primarily infer structure from text, whereas group chats expose native structural signals, including reply links, participant roles, and cross-channel membership, that are not fully captured by text alone.

\textbf{Multi-party conversation understanding} spans three directions: 
inferring participant states (e.g., emotion recognition~\cite{ghosal-etal-2019-dialoguegcn}), modeling discourse structure (e.g., conversation disentanglement~\cite{kummerfeld-etal-2019-large}), and summarizing group interactions 
(e.g., dialogue summarization~\cite{gliwa-etal-2019-samsum}, decision  detection~\cite{addlesee-etal-2023-multi}). 
Recent work couples group chats with LLM agents for ongoing participation, for example by jointly generating responses and predicting the next speaker~\cite{wang2025multi} or orchestrating multi-party tool use~\cite{houde2025controlling}.
In contrast, we treat completed group chats as a knowledge source and organize them into structured, evidence-grounded memory for downstream agents.

\section{Methodology}
\label{sec:method}

\model{} organizes enterprise group chats into three complementary forests over a shared, immutable message log, preserving structural signals lost when chats are compressed into summaries or treated as independent chunks (Figure~\ref{fig:framework}).
Offline, it constructs a \emph{Channel forest} for temporal workflows, an \emph{Actor forest} for speaker-conditioned context, and an \emph{Artifact forest} for cross-channel deliverable lifecycles.
Online, four stages retrieve message anchors, supplement them through summaries, recover linked evidence, and assemble a token-bounded context for the answer agent.

\subsection{Structured Memory Construction}
\label{sec:memory-tree}

\subsubsection{Three-View Forest Architecture}
\label{sec:forests}

\paragraph{Immutable Message Log.}
The three forests share an immutable log of verbatim messages. Each forest leaf references a single log entry rather than copying its content, preserving traceability to the original evidence and avoiding duplication across views.
For each message $m$, the log stores its timestamp $t_m$, sender $u_m$, channel $c_m$, optional reply target $r_m$, and raw text $x_m$. 
We further derive a mention set $\mathrm{mentions}(m) \subseteq \mathcal{U}$ and structured identifiers, such as document codes, from $x_m$ using lightweight regular expressions.

\paragraph{Channel forest.}
The Channel forest organizes each channel as a temporal hierarchy, making conversational context retrievable at multiple temporal scales.
For each channel $c \in \mathcal{C}$, we construct a balanced $k$-ary tree $T_c$ whose leaves are references to the immutable log entries of messages in $c$, ordered chronologically.
Every $k$ consecutive leaves or child nodes are recursively grouped into a parent until a single root remains.
Each internal node summarizes its contiguous time interval, focusing on participants, workflows, decisions, commitments, blockers, and unresolved issues.
Unlike semantic clustering hierarchies, this construction preserves temporal contiguity and event order, supporting retrieval of project progress and decision evolution.
The running example summarizes the \texttt{\#backend} workflow from merge to release, preserving traceability to message-level evidence, including the QA requirement that governs the transition.

\methodexample{}{%
{\scriptsize\ttfamily
/* In all running examples, [R] denotes a root node,
   [N] an internal node with an LLM-generated summary, and
   [L] a leaf node that references an immutable log entry. */
\par\smallskip
}
{\fontsize{8}{9}\selectfont
\begin{tabular}{@{}l@{}}
\texttt{\textbf{[R]} \#backend [channel]} \\
\texttt{+-- \textbf{[N]} Dec 1: merge and release gate} \\
\texttt{|~+-- \textbf{[L]} $m_{091}$: @Bob, PR for DOC-042 is merged.} \\
\texttt{|~+-- \textbf{[L]} $m_{092}$: Staging looks good.} \\
\texttt{|~+-- \textbf{[L]} $m_{093}$: Lunch at 12?} \\
\texttt{|~+-- \textbf{[L]} $m_{094}$: Need QA sign-off before release.} \\
\texttt{+-- \textbf{[N]} Dec 3: release confirmation} \\
\texttt{~~+-- \textbf{[L]} $m_{099}$: Released after Bob's approval.}
\end{tabular}
}
}

\paragraph{Actor forest.}
The Actor forest addresses speaker-conditioned semantics by organizing messages around each participant, enabling cross-channel tracing of an employee's responsibilities, commitments, interactions, and reported context.
For each participant $u \in \mathcal{U}$, we construct an actor tree $T_u$ whose root separates messages authored by $u$ from messages that explicitly mention $u$.
Within each branch, messages are partitioned by channel, ordered chronologically, and recursively grouped using the same $k$-ary construction as the Channel forest.
Internal-node summaries emphasize the participant's speech acts, responsibilities, expertise, ownership, and reporting relations.
The running example aggregates Bob's authored and mentioned messages, forming a participant-centered record of his project involvement, responsibilities, and interaction patterns.

\methodexample{}{%
{\fontsize{8}{9}\selectfont
\begin{tabular}{@{}l@{}}
\texttt{\textbf{[R]} Bob [actor]} \\
\texttt{+-- \textbf{[N]} authored-by} \\
\texttt{|~+-- \textbf{[N]} \#backend} \\
\texttt{|~|~+-- \textbf{[N]} Dec 1: release requirement} \\
\texttt{|~|~~~+-- \textbf{[L]} $m_{094}$: Need QA sign-off before release.} \\
\texttt{|~+-- \textbf{[N]} \#release} \\
\texttt{|~~~+-- \textbf{[N]} Dec 3: release approval} \\
\texttt{|~~~~~+-- \textbf{[L]} $m_{098}$: Approved. Release DOC-042 today.} \\
\texttt{+-- \textbf{[N]} mentioned-in} \\
\texttt{~~+-- \textbf{[N]} \#backend} \\
\texttt{~~~~+-- \textbf{[N]} Dec 1--3: merged and released} \\
\texttt{~~~~~~+-- \textbf{[L]} $m_{091}$: @Bob, PR for DOC-042 is merged.} \\
\texttt{~~~~~~+-- \textbf{[L]} $m_{099}$: Released after Bob's approval.}
\end{tabular}
}
}

\paragraph{Artifact forest.}
The Artifact forest organizes messages around enterprise deliverables, enabling cross-channel tracing of an artifact's lifecycle, status, and ownership.
We identify artifacts offline by clustering messages that discuss the same artifact. 
We extract artifact-like identifiers, such as document codes, from messages using lightweight rules and cluster the associated messages using farthest-first clustering.
Candidate clusters with highly similar centroids are merged automatically, while an LLM judge resolves borderline cases. We retain clusters with sufficient lifecycle transitions.
For each retained artifact $a$, we construct a tree $T_a$ whose leaves reference its member messages pooled across channels and ordered chronologically.
Messages are recursively grouped using the same $k$-ary construction as the other forests. Internal-node summaries emphasize lifecycle transitions, responsible participants, and cross-channel status changes.
The running example links merge, QA, approval, and release evidence from two channels into a traceable lifecycle for DOC-042.

\methodexample{}{%
{\fontsize{8}{9}\selectfont
\begin{tabular}{@{}l@{}}
\texttt{\textbf{[R]} DOC-042 [artifact]} \\
\texttt{+-- \textbf{[N]} Dec 1: merged and awaiting QA} \\
\texttt{|~+-- \textbf{[L]} $m_{091}$: @Bob, PR for DOC-042 is merged.} \\
\texttt{|~+-- \textbf{[L]} $m_{094}$: Need QA sign-off before release.} \\
\texttt{+-- \textbf{[N]} Dec 3: QA passed, approved, released} \\
\texttt{~~+-- \textbf{[L]} $m_{095}$: @Lily PR Check DOC-042.} \\
\texttt{~~+-- \textbf{[L]} $m_{097}$: DOC-042 passed QA.} \\
\texttt{~~+-- \textbf{[L]} $m_{098}$: Approved. Release DOC-042 today.} \\
\texttt{~~+-- \textbf{[L]} $m_{099}$: Released after Bob's approval.}
\end{tabular}
}
}

\subsubsection{Cross-View Connections}
\label{sec:cross-view}

To surface cross-view evidence efficiently at query time, we build cross-view connections through a \emph{Placement Map} and a sparse set of \emph{Bridges}.

\paragraph{Placement Map.}
Each forest indexes references to messages in the same immutable message log under a different routing rule. Consequently, a message may have placements in multiple trees across views, providing a direct structural connection between them. To make this link efficiently queryable, we precompute for every message $m$ the placement table
$\mathrm{Pl}(m)=\{(\text{view}, \text{tree\_id}, \text{parent\_id}, \text{leaf\_pos})\}$
indexed by $\text{msg\_id}$ in a hash map. The Placement Map supports three structural retrieval primitives through direct lookup rather than tree traversal. $\mathrm{trees\_of}(m)$ returns the trees containing $m$. $\mathrm{window}(m,T,w)$ returns the $w$ messages adjacent to $m$ in the leaf order of tree $T$. $\mathrm{cotenants}(m)$ returns the sibling leaves that share a parent node with $m$ across its placements. These primitives make cross-view placements and local structural neighborhoods directly addressable, thereby supporting the multi-view joins used later in retrieval.

\paragraph{Bridges.}
Whereas the Placement Map connects multiple placements of the same message, bridges connect distinct messages through strong structural or evidential relations. We construct two bridge types from native reply links and the leaf-level identifiers defined in Section~\ref{sec:forests}. \emph{Reply bridges} connect a message $m$ with its explicit reply target $r_m$ when $r_m \neq \emptyset$. \emph{Identifier bridges} connect messages sharing a structured identifier, such as a document code or ticket ID. To avoid dense all-pairs connectivity, each message links to at most $b$ same-identifier messages ranked by temporal proximity.
These sparse bridges connect local conversational dependencies and cross-channel artifact evidence, supporting the bridge-hop operator in retrieval.

\subsection{Forest Retrieval}
\label{sec:tree-retrieval}

Given a query $q$, retrieval produces a compact evidence set $\mathcal{E}_q$ under a fixed token budget through four stages. $S_1$ retrieves primary evidence via view-aware leaf recall, $S_2$ supplements it with summary-guided nodes, $S_3$ completes the chain through bounded cross-view operations, and $S_4$ assembles the final context. 
Before $S_1$, we parse $q$ into a lightweight query specification using
deterministic rules that require no LLM calls.
The specification records the asker and mentioned entities, including
participants, channels, and artifact identifiers, to guide subsequent view
selection.
We detail $S_1$ through $S_4$ below.

\subsubsection{$S_1$ View-aware Leaf Retrieval}
\label{sec:s1}

$S_1$ combines global message seeding with view-aware tree routing. Global
retrieval identifies fine-grained message-level anchors, while tree routing
lifts these anchors to coarser structural contexts that capture broader
workflows, participant histories, and artifact lifecycles. The selected
trees define the search space for subsequent summary expansion and
cross-view completion.

\paragraph{Candidate seeding.}
Given a query $q$, we retrieve messages using BM25 for lexical matching and
dense retrieval for semantic matching. We fuse the resulting rankings through
reciprocal rank fusion (RRF), which combines ranks rather than raw scores
with incompatible scales:
\begin{equation}
\small
s_{\mathrm{hyb}}(m,q)
=
\frac{1}{\eta+\mathrm{rank}_{\mathrm{bm25}}(m)}
+
\lambda_d
\cdot
\frac{1}{\eta+\mathrm{rank}_{\mathrm{den}}(m)},
\end{equation}
where $\eta$ is a smoothing constant and $\lambda_d$ weights dense retrieval. 
The top-$K$ messages under $s_{\mathrm{hyb}}(m,q)$ form the anchor pool. 
Rather than being returned directly, these anchors are projected to their
forest placements to identify relevant trees and select leaf evidence within
each view.

\paragraph{View-aware routing.}
Using the anchor pool, we score each candidate tree with three
complementary signals. \emph{Leaf2tree} projects each anchor to its host
trees through $\mathrm{Pl}(\cdot)$ and aggregates weighted anchor votes.
\emph{Root} matches $q$ against each tree's root summary.
\emph{Collapsed} captures locally relevant content that may be obscured by
a tree's coarse root summary. It scans internal-node summaries by cosine
similarity and assigns each tree the score of its best-matching summary.
We rank each signal within each view rather than globally, and fuse the
resulting within-view ranks using RRF. 
We select top-ranked trees for subsequent stages and retain anchor messages
as primary leaf evidence under fixed per-view quotas, preventing larger
views from crowding out complementary ones.

\subsubsection{$S_2$ Summary-guided Supplementation}
\label{sec:s2}

$S_2$ supplements the message-level evidence returned by $S_1$ with
summary-level retrieval over the trees selected in $S_1$.
A message may weakly match $q$ in isolation even when its surrounding segment is highly relevant. We therefore score internal-node summaries against $q$, recovering such messages and providing layered context for the answer model.
For each internal node $n$ with summary embedding $\mathbf{s}_n$, we compute
$\cos(\mathbf{q}, \mathbf{s}_n)$. To avoid overlap, we retain
only the highest-scoring node on each root-to-leaf path and expand it into a
bounded number of descendant leaves. These leaves are appended with their
ancestor summary chain as \emph{navigation headers}. We place $S_2$ leaves
after $S_1$ messages rather than merging raw scores, as RRF and cosine
similarity are not directly comparable.

\subsubsection{$S_3$ Structural Completion}
\label{sec:s3}

$S_3$ recovers cross-view evidence via bounded Placement Map joins and bridge hops.

\paragraph{Placement Map join.}
To avoid unnecessary expansion, $S_3$ invokes a Placement Map join only when
the query mentions at least two resolvable entities. It first locates an
evidence message $m_A$ for one entity $A$ from $S_1$, then uses
$\mathrm{Pl}(m_A)$ to enter a related placement in the selected view.
For queries with an explicit temporal cue, candidates are restricted to
leaves occurring before or after $m_A$.
Candidates are ranked by similarity to the remaining entity $B$ and $q$,
and the top candidate is retained only if its score exceeds a threshold.

\paragraph{Bridge hop.}
$S_3$ expands high-confidence leaves from $S_1$ and $S_2$ through the sparse
bridges defined in Section~\ref{sec:cross-view}. 
For each candidate leaf $m$, we inspect its incident reply and identifier bridges, and add each reached message $m'$ to the candidate pool as a bridge anchor. 
For each $m$, $S_3$ follows at most one bridge into a tree outside the evidence pool,
choosing the endpoint most relevant to $q$. 
We then inspect a bounded window around the selected endpoint in that tree and adds the
most query-relevant neighboring leaf.
A small hop budget prevents uncontrolled expansion while recovering reply
dependencies and cross-channel artifact evidence.

\subsubsection{$S_4$ Budget-aware Assembly}
\label{sec:s4}

$S_4$ merges and deduplicates candidates from $S_1$, $S_2$, and $S_3$.
Since scores are not comparable across stages, we preserve within-stage ranks and add candidates in the order $S_1$, $S_2$, and $S_3$.
Each stage is subject to a fixed allocation, and selection stops when the token budget is
reached.
Selected messages are rendered verbatim with lightweight metadata (channel, sender, lifecycle markers). 
Summary headers from $S_2$ are attributed to their originating nodes and capped at a
small fraction of the budget, allowing the answer model to distinguish raw
evidence from generated navigation. 
This design preserves traceability to the original chat log.


\section{Experiments}
\label{sec:experiment}

We evaluate \model{} through four key questions:
\textbf{Q1 (Section~\ref{sec:main_results}):} Does \model{} outperform
RAG and memory-system baselines?
\textbf{Q2 (Section~\ref{sec:component_study}):} Do the Channel, Actor,
and Artifact views provide complementary benefits?
\textbf{Q3 (Section~\ref{sec:retrieval_ablation}):} Does \model{} retrieve more relevant evidence than RAG baselines, and how does each retrieval stage ($S_1$--$S_4$) contribute?
\textbf{Q4 (Section~\ref{sec: efficiency}):} How efficient is \model{} in
terms of offline construction cost, storage footprint, and online retrieval latency?

\subsection{Experimental Setup}
\label{sec:setup}

\paragraph{Datasets.}
We employ \textsf{EverMemBench}~\citep{EverMemBench2026},
\textsf{GroupMemBench}~\citep{yang2026groupmembench}, and a real-world internal enterprise benchmark \textsf{Company}. 
Details are in Appendix~\ref{sec:dataset_details}.

\paragraph{Baselines.}
RAG baselines include 
\texttt{BM25}~\citep{robertson2009probabilistic},
\texttt{Dense}~\citep{karpukhin2020dense},
their reciprocal-rank-fusion variant
\texttt{BM25+Dense}~\citep{cormack2009reciprocal}, and
\texttt{GraphRAG}~\citep{edge2024graphrag}.
Agent memory-system baselines include
\texttt{MemoBase}~\citep{team2025memobase},
\texttt{Mem0}~\citep{chhikara2025mem0},
\texttt{A-Mem}~\citep{xu2026mem}, and
\texttt{MemOS}~\citep{li2025memos}.

\paragraph{Metrics.}
We report \textbf{Answer Accuracy}, measured against gold answers LLMs. 
For \textsf{Company}, we additionally report
\textbf{Precision@$k$} and \textbf{Recall@$k$} against gold evidence messages.

\paragraph{Implementation.}
Following the official protocol, we use GPT-5 as the answer model on \textsf{GroupMemBench} and GPT-4.1-mini on \textsf{EverMemBench}. We also use GPT-4.1-mini on \textsf{Company}.
We use the frozen \texttt{Qwen3-Embedding-0.6B} encoder, set the branching factor to $k=16$, and retrieve under a 3,000-token budget. Details are in Appendix~\ref{sec:hyperparam}.

\subsection{Q1: Main Results}
\label{sec:main_results}

\begin{table*}[!t]
\centering
\setlength{\tabcolsep}{3.6pt}
\renewcommand{\arraystretch}{1.08}

\datasettitle{EverMemBench}

\begingroup
\setlength{\aboverulesep}{0.3ex}
\setlength{\belowrulesep}{0.3ex}

\resizebox{\textwidth}{!}{
\tiny
\renewcommand{\arraystretch}{0.5}
\begin{tabular}{lcccccccccc}
\specialrule{0.11em}{0.8ex}{0.8ex}
\multirow{2}{*}{Method}
& \multicolumn{3}{c}{Fine-grained Recall}
& \multicolumn{3}{c}{Memory Awareness}
& \multicolumn{3}{c}{Profile Understanding}
& \multirow{2}{*}{Avg.} \\

\cmidrule(lr){2-4}
\cmidrule(lr){5-7}
\cmidrule(lr){8-10}
& Single & Multi & Temp.
& Const. & Proact. & Update
& Style & Skill & Role
& \\
\specialrule{0.06em}{0.8ex}{0.8ex}

BM25
& 87.45 & 5.62 & 11.33
& 70.60 & 62.20 & 37.28
& 30.94 & 43.44 & 53.63
& 44.72 \\

Dense
& 86.23 & 6.83 & 14.33
& \underline{79.60} & \textbf{70.20} & \underline{59.93}
& 31.49 & \underline{45.70} & 55.24
& \underline{49.95} \\

BM25+Dense
& 81.38 & 4.42 & 14.00
& 75.20 & \underline{68.40} & 49.48
& 32.60 & \textbf{46.61} & \textbf{57.26}
& 47.70 \\

GraphRAG
& 69.64 & 5.22 & 14.33
& 79.00 & 49.00 & 52.96
& \textbf{38.12} & \underline{45.70} & 50.00
& 44.89 \\

\specialrule{0.06em}{0.8ex}{0.8ex}

MemoBase
& 60.09 & 12.85 & \underline{18.00}
& 64.68 & 36.77 & 30.60
& 17.05 & 29.59 & 38.78
& 34.27 \\

Mem0
& 55.40 & 11.24 & 6.33
& 66.17 & 52.46 & 51.87
& 22.73 & 31.36 & 36.22
& 37.09 \\

A-Mem
& \underline{88.26} & 6.83 & 11.67
& 77.20 & 62.20 & 48.78
& \underline{34.81} & \textbf{46.61} & 50.00
& 47.37 \\

MemOS
& 71.36 & \textbf{18.88} & 15.67
& 69.90 & 51.99 & 45.15
& 28.98 & 32.54 & 48.47
& 42.55 \\

\specialrule{0.06em}{0.8ex}{0.8ex}

\textsc{Ours}
& \textbf{89.88} & \underline{17.27} & \textbf{33.33}
& \textbf{81.00} & \textbf{70.20} & \textbf{61.32}
& \textbf{38.12} & \underline{45.70} & \underline{55.65}
& \textbf{54.72} \\
\specialrule{0.11em}{0.8ex}{0.8ex}
\end{tabular}
}
\endgroup


\datasettitle{GroupMemBench}

\resizebox{\textwidth}{!}{
\fontsize{4.5pt}{4.5pt}\selectfont
\renewcommand{\arraystretch}{0.3}
\begin{tabular}{lccccccc}
\specialrule{0.1em}{0.8ex}{0.8ex}
Method
& Multi-Hop
& Update
& Ambiguity
& Implicit
& Temporal
& Abstention
& Avg. \\
\specialrule{0.06em}{0.8ex}{0.8ex}

BM25
& \underline{40.66} & \underline{28.04} & 16.04 & \underline{46.94} & \underline{56.17} & \underline{83.45} & \underline{47.11} \\

Dense
& 35.16 & 24.30 & 22.64 & 44.90 & 43.83 & 77.70 & 42.28 \\

BM25+Dense
& 32.42 & 22.43 & 23.58 & 30.61 & 43.83 & \textbf{84.17} & 41.74 \\

GraphRAG
& 17.03 & 19.63 & 17.92 & 20.41 & 6.79 & 76.26 & 26.58 \\

\specialrule{0.06em}{0.8ex}{0.8ex}

MemoBase
& 15.93 & 3.74 & 15.09 & 20.41 & 20.99 & 82.73 & 27.92 \\

Mem0
& 23.63 & 14.02 & 16.04 & 24.49 & 7.41 & 79.86 & 28.19 \\

A-Mem
& 31.87 & 27.10 & \underline{26.42} & 36.73 & 21.60 & 78.42 & 37.18 \\

MemOS
& 28.57 & 21.50 & 24.53 & 18.37 & 25.93 & 82.01 & 35.70 \\

\specialrule{0.06em}{0.8ex}{0.8ex}

\textsc{Ours}
& \textbf{47.25} & \textbf{32.71} & \textbf{40.57} & \textbf{55.10} & \textbf{68.52} & 81.29 & \textbf{55.70} \\

\specialrule{0.1em}{0.8ex}{0.8ex}
\end{tabular}
}

\datasettitle{Company Dataset}

\resizebox{\textwidth}{!}{
\scriptsize
\renewcommand{\arraystretch}{0.85}
\begin{tabular}{lcccccccccccc}
\toprule
\multirow{2}{*}{Method}
& \multicolumn{3}{c}{Fine-grained Recall}
& \multicolumn{4}{c}{Memory Awareness}
& \multicolumn{4}{c}{Profile Understanding}
& \multirow{2}{*}{Avg.} \\
\cmidrule(lr){2-4}
\cmidrule(lr){5-8}
\cmidrule(lr){9-12}
& Single & Multi-hop & Temp.
& Const. & Update & Implicit & Abstention
& Style & Skill & Role & Ambiguity
& \\
\midrule

BM25
& 58.82 & 44.87 & 41.27
& \underline{75.00} & 42.19 & 45.05 & 50.00
& 2.13 & 8.33 & 5.00 & 75.00
& 37.45 \\

Dense
& 57.98 & 48.72 & \underline{47.62}
& 66.25 & 35.94 & 45.05 & \underline{62.50}
& 1.06 & 7.14 & 2.50 & \underline{80.00}
& 36.58 \\

BM25+Dense
& \underline{62.18} & 53.85 & 44.44
& 72.50 & \underline{48.44} & \underline{55.86} & 37.50
& \underline{4.26} & 5.95 & 2.50 & \textbf{85.00}
& \underline{40.70} \\

GraphRAG
& 56.30 & \underline{55.13} & 33.33
& 57.50 & 34.38 & 48.65 & 37.50
& \underline{4.26} & 5.95 & \underline{7.50} & 75.00
& 35.71 \\

\midrule

MemoBase
& 36.97 & 34.62 & 23.81
& 37.50 & 29.69 & 34.23 & 12.50
& 0.00 & 5.95 & 3.75 & 60.00
& 24.22 \\

Mem0
& 52.94 & 39.74 & 33.33
& 63.75 & 34.38 & 42.34 & 37.50
& 2.13 & \underline{10.71} & 5.00 & 70.00
& 33.33 \\

A-Mem
& 59.66 & 52.56 & \underline{47.62}
& 67.50 & 39.06 & 48.65 & \underline{62.50}
& 3.19 & 7.14 & 6.25 & 75.00
& 38.58 \\

MemOS
& 56.30 & 33.33 & 36.51
& 70.00 & 29.69 & 42.34 & 12.50
& 1.06 & 5.95 & 2.50 & 65.00
& 32.46 \\

\midrule

\textsc{Ours}
& \textbf{78.15} & \textbf{65.38} & \textbf{61.90}
& \textbf{86.25} & \textbf{64.06} & \textbf{73.87} & \textbf{75.00}
& \textbf{7.45} & \textbf{14.29} & \textbf{12.50} & \textbf{85.00}
& \textbf{53.31} \\
\bottomrule
\end{tabular}
}

\caption{
Main results on three enterprise group benchmarks.
We use the official taxonomies for \textsf{EverMemBench} and \textsf{GroupMemBench}, and a unified
\texttt{group\_chat} taxonomy for \textsf{Company}.
Bold: best; underlined: second-best.
}
\label{tab:main_three_datasets}

\end{table*}

\paragraph{Overall.}
Table~\ref{tab:main_three_datasets} shows that \model{} achieves the highest average accuracy on all three benchmarks, exceeding the strongest baseline by $4.77$, $8.59$, and $12.61$ percentage points, respectively.
These consistent gains demonstrate its robustness across diverse dialogue settings and question types.

\paragraph{Dataset and category analysis.}
\model{} ranks first or second on most subcategories.
On \textsf{EverMemBench}, which contains long-horizon, multi-channel project discussions, \model{} performs especially well on Fine-grained Recall and Memory Awareness, which require \textbf{linking evidence across channels and time}. 
Profile Understanding remains difficult because Style, Skill, and Role must be inferred from diffuse behavioral patterns rather than explicit event-level evidence.
On \textsf{GroupMemBench}, which emphasizes threaded conversations and asker-conditioned queries, \model{} achieves $55.10$ on Implicit and $40.57$ on Ambiguity. 
Implicit questions require tracing first-person references through the asker's history, whereas Ambiguity questions require resolving terminology across speakers.
These results are consistent with the benefit of the Actor forest for retrieving \textbf{asker-conditioned context}.
\model{} is less dominant on Abstention ($81.29$), where recall-oriented retrieval may surface plausible but non-supporting messages.
On \textsf{Company}, \model{} reaches $61.90$ on Temporal and $64.06$ on Update, where the Channel forest preserves workflow chronology and the Artifact forest links lifecycle evidence across channels.

\paragraph{Understanding the baseline gap.}
Memory systems may merge claims across speakers or threads when consolidating messages into facts, profiles, or summaries, obscuring attribution, updates, and verbatim evidence. On \textsf{Company}, \texttt{MemOS} obtains only 33.33/36.51 on Multi-hop/Temporal, versus 65.38/61.90 for \model{}.
Raw-message RAG preserves message details but ranks messages independently.
On \textsf{Company}, \texttt{BM25+Dense} obtains 53.85/44.44 on Multi-hop/ Temporal questions requiring evidence across channels and time.
GraphRAG further abstracts messages into entity-relation graphs and community summaries, diluting reply, attribution, and workflow signals.

\subsection{Q2: Component Study for \model{}}
\label{sec:component_study}

\begin{figure}[!t]
\centering
\includegraphics[width=\linewidth]{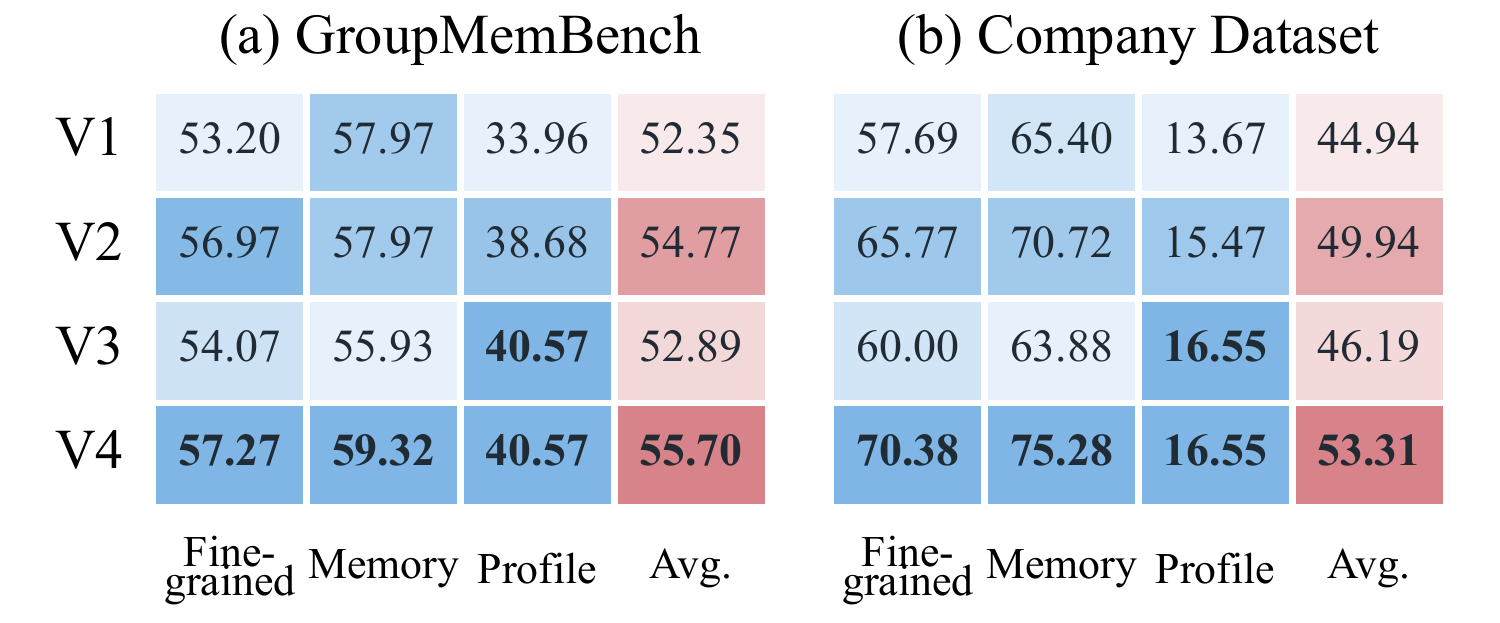}
\caption{Performance of different view combinations on \textsf{GroupMemBench} and \textsf{Company}.
V1: Channel; V2: Channel+Actor; V3: Channel+Artifact; V4: Full.}

\label{fig:view_heatmap}
\end{figure}

Figure~\ref{fig:view_heatmap} evaluates the contribution of the three
views\footnote{On GroupMemBench, Fine-grained Recall (Multi-hop + Temporal),Memory Awareness (Update + Implicit + Abstention), and Profile Understanding(Ambiguity).}.
The full model (V4) achieves the best weighted average on both datasets.
Compared with V1, V2 improves Fine-grained Recall and Memory Awareness on the Company dataset by 8.08 and 5.32 points, respectively, showing the benefit of speaker-centered context. 
Compared with V1, V3 improves Fine-grained Recall by 0.87/2.31 points and Profile Understanding by 6.61/2.88 points on GroupMemBench/Company,
respectively. 
By grouping messages about the same deliverable, the Artifact view exposes cross-channel lifecycle and ownership evidence useful for evidence recall and profile inference. 
Finally, V4 outperforms V2 by 0.93 points on GroupMemBench and 3.37 points on the Company dataset, confirming that Artifact evidence complements the Channel and Actor views.

\subsection{Q3: Retrieval Performance}
\label{sec:retrieval_ablation}

\subsubsection{Comparison with Standard Retrieval}
\model{} achieves the strongest overall retrieval performance, attaining the highest Precision@$k$ at every budget while maintaining competitive Recall@$k$\footnote{Precision@$k$ is inherently low: with 3.6 gold messages per question,P@20 is upper-bounded at roughly 18\%.}.
This result is consistent with its multi-view organization and structural retrieval pipeline, which consider temporal, participant, and artifact context in addition to message-level relevance.
GraphRAG obtains the highest Recall@20 but at substantially lower precision, suggesting noisier expansion. 
Lexical and dense baselines improve more slowly as the
retrieval budget increases, likely because they rank messages independently and cannot exploit multi-view structural relations.

\begin{figure}[!htbp]
\centering
\includegraphics[width=\linewidth]{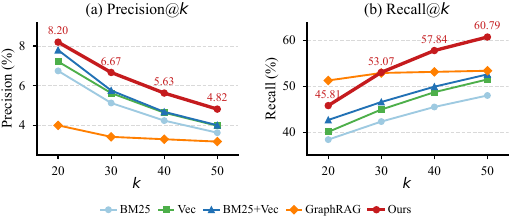}
\caption{Retrieval Performance on \textsf{Company}.}
\label{fig:retrieval_pr}
\end{figure}

\subsubsection{Stage-wise Contribution}
Table~\ref{tab:retrieval_ablation} shows the cumulative contribution of each retrieval stage under the same context budget.
S1 reaches 40.3 R@20, suggesting that hybrid message anchors and routing through complementary Channel, Actor, and Artifact views recover much of the directly relevant evidence.
S2 adds summary-guided supplementation and yields modest gains in R@20 and accuracy ($40.3\!\to\!40.8$ and $45.4\!\to\!45.8$), indicating that S1 covers much of the directly relevant evidence. 
S3 improves P@20 from 7.1 to 8.0 and R@20 from 40.8 to 44.5 through Placement Map joins and bridge hops to recover evidence distributed across views. 
S4 yields small retrieval gains but improves accuracy from 47.9 to 53.3, showing that effective evidence selection and organization are crucial for downstream answering.

\begin{table}[!htbp]
\centering
\resizebox{\columnwidth}{!}{
\renewcommand{\arraystretch}{0.85}
\begin{tabular}{lccc}
\toprule
Variant & P@20 & R@20 & Acc. \\
\midrule

S1: View-aware Leaf Retrieval
& 7.0 & 40.3 & 45.4 \\

+ S2: Summary-guided Supplementation
& 7.1 & 40.8 & 45.8 \\

+ S3: Structural Completion
& 8.0 & 44.5 & 47.9 \\

\rowcolor{gray!12}
\textbf{Full (+ S4: Budget-aware Assembly)}
& \textbf{8.2} & \textbf{45.8} & \textbf{53.3} \\

\bottomrule
\end{tabular}
}
\caption{Stage-wise retrieval ablation of \model{}.}
\label{tab:retrieval_ablation}
\end{table}

\subsection{Q4: Efficiency}
\label{sec: efficiency}

\model{} incurs moderate offline cost. Its three-forest construction is slower
than lexical and dense indexing but faster than GraphRAG, MemoBase, and Mem0,
with a 206\,MB storage footprint. At query time, \model{} achieves the fastest
retrieval latency at 65\,ms per query, enabled by cached node summaries and
constant-time Placement Map lookups. Its answer-token usage is comparable to
other memory systems, averaging 2,365 input and 1,300 output tokens per query.

\begin{table}[!htbp]
\centering
\resizebox{\columnwidth}{!}{
\renewcommand{\arraystretch}{0.85}
\begin{tabular}{
l
>{\columncolor{timeblue}}c
c
c
>{\columncolor{timeblue}}c
c
>{\columncolor{tokengreen}}c
}
\toprule
\multirow{2}{*}{Method}
& \multicolumn{2}{c}{Offline Construction}
&
& \multicolumn{1}{c}{Online Retrieval}
&
& \multicolumn{1}{c}{Online Answer} \\
\cmidrule(lr){2-3}
\cmidrule(lr){5-5}
\cmidrule(lr){7-7}
& Build Time
& Storage
&
& Ret. Time / Q
&
& Answer Tok. / Q \\
\midrule
BM25        & 0.6 s & -- & & 253 ms & & 1454 / 103 \\
Vec         & 3.4 s & 83 MB & & 72 ms & & 1462 / 112 \\
BM25 + Vec  & 8.9 s & 83 MB & & 710 ms & & 1452 / 106 \\
GraphRAG & 38200 s & 641 MB & & 2107 ms & & 3936 / 1283 \\
 \midrule
MemoBase    & 7994 s & -- & & 1634 ms & & 3153 / 1422 \\
Mem0        & 7706 s & 1.4 GB & & 15339 ms & & 492 / 1249 \\
A-Mem       & 434 s & 24 MB & & 97 ms & & 2305 / 1311 \\
MemOS       & 1401 s & -- & & 573 ms & & 632 / 1207 \\ \midrule
Ours        & 1842 s & 206 MB & & 65 ms & & 2365 / 1300 \\
\bottomrule
\end{tabular}
}
\caption{Efficiency comparison on GroupMemBench.}
\label{tab:cost_token_colored}
\end{table}

\section{Conclusion}
\label{sec:conclusion}
As an initial exploration of agent memory for enterprise group chats, \model{} organizes immutable message evidence through complementary Channel, Actor, and Artifact forests. This structured-memory paradigm enables agents to navigate traceable organizational evidence across time, participants, and artifacts. Experiments on three benchmarks demonstrate consistent improvements across diverse question categories.

\section*{Limitations}

\model{} is evaluated through question answering on three enterprise-chat benchmarks. This setting enables controlled comparison of memory quality, while leaving its benefits for broader agent workflows, such as task coordination and long-term planning, to be explored. Moreover, \model{}'s offline construction time remains manageable at the evaluated scale, making periodic memory updates through full reconstruction feasible. For larger or more frequently updated workspaces, incremental forest construction and selective updates to summaries and cross-view connections are promising directions for balancing memory freshness with maintenance cost. Finally, our current formulation focuses on text messages and their associated metadata. Images, document attachments, and voice messages can contain relevant organizational knowledge beyond the message text. Future work could extend \model{} to integrate these sources into a unified multimodal memory while preserving traceability to the original evidence.

\bibliography{custom}

\clearpage
\appendix

\section{Appendix}

\subsection{Additional Experiments}
\label{sec:dataset_details}

\subsubsection{Answer-Model Robustness}
\label{sec: agent_backbone}

\model{} consistently outperforms memory-system baselines across all answer-model backbones, with gains of 13.83--18.52 points over the strongest baseline. 
While baseline rankings vary across backbones, \model{} remains the top-performing method. 
It achieves 57.18\% with GLM-5.3 and 55.70\% with GPT-5, showing that its structured evidence remains effective across answer models.

\begin{table}[!htbp]
\centering
\scriptsize
\setlength{\tabcolsep}{2.8pt}
\renewcommand{\arraystretch}{1.12}

\resizebox{\columnwidth}{!}{%
\begin{tabular}{lccccc}
\toprule
\textbf{Backbone}
& \textbf{MemoBase}
& \textbf{Mem0}
& \textbf{MemOS}
& \textbf{A-Mem}
& \textbf{Ours} \\
\midrule

Qwen3.5-35B-A3B
& 25.23 & 24.7 & 32.08 & 33.69 & \textbf{47.52} \\

DeepSeek-V4-Flash
& 28.46 & 26.44 & 32.75 & 37.99 & \textbf{52.35} \\

GLM-5.3
& 31.01 & 29.66 & 35.44 & 40.54 & \textbf{57.18} \\

\midrule

GPT-5 (default)
& 27.92 & 28.19 & 35.70 & 37.18 & \textbf{55.7} \\

\bottomrule
\end{tabular}%
}
\caption{
Agent backbone ablation on GroupMemBench.
}
\label{tab:backbone_groupmembench}
\end{table}

\subsection{Dataset Details}
\label{sec:dataset_details}

We evaluate \model{} on two public enterprise group-chat memory
benchmarks and an internal enterprise-chat benchmark. Table
\ref{tab:evermembench_data}, Table~\ref{tab:groupmembench_stats}, and
Table~\ref{tab:internal_data} summarize their scales and evaluation
distributions.

\subsubsection{EverMemBench}

\textsc{EverMemBench}~\citep{EverMemBench2026} evaluates long-horizon
collaborative memory over nterprise group chats spanning multiple channels.
We use its official evaluation protocol and all released question-answer
pairs.

\begin{table}[!htbp]
\centering
\small
\resizebox{\columnwidth}{!}{
\begin{tabular}{llr}
\toprule
\textbf{Category} & \textbf{Statistic} & \textbf{Value} \\
\midrule
Corpus & Projects & 5 \\
& Channels per project & 3 \\
& Participants & 170 \\
& Time span & 365 days \\
& Messages & 51,023 \\
& Dialogue tokens & 4.23M \\
\midrule
Fine-grained Recall & Single-hop / Multi-hop / Temporal & 213 / 249 / 300 \\
Memory Awareness & Constraint / Proactivity / Update & 402 / 427 / 268 \\
Profile Understanding & Style / Skill / Role & 176 / 169 / 196 \\
\midrule
Evaluation & Total QA pairs & 2,400 \\
\bottomrule
\end{tabular}
}
\caption{Statistics and question distribution of \textsc{EverMemBench}.}
\label{tab:evermembench_data}
\end{table}

\subsubsection{GroupMemBench}
\textsc{GroupMemBench}~\citep{yang2026groupmembench} evaluates memory over
threaded enterprise group chats with user-conditioned queries. We use
the official corpus and evaluation protocol.

\begin{table}[t]
\centering
\small
\begin{tabular}{lll}
\toprule
\textbf{Category} & \textbf{Statistic} & \textbf{Value} \\
\midrule
\multirow{6}{*}{Corpus}
& Domains & 4 \\
& Projects & 33 \\
& Topics & 125 \\
& Phases & 295 \\
& Participants & 45 \\
& Messages & 120,000 \\
\midrule
\multirow{4}{*}{Message Type}
& Posts & 27,222 \\
& Threaded replies & 79,766 \\
& Cross-project replies & 13,012 \\
& Decision changes & 352 \\
\midrule
\multirow{7}{*}{Evaluation}
& Multi-Hop & 182 \\
& Update & 107 \\
& Ambiguity & 106 \\
& Implicit & 49 \\
& Temporal & 162 \\
& Abstention & 139 \\
& Total QA pairs & 745 \\
\bottomrule
\end{tabular}
\caption{Statistics and evaluation-question distribution of GroupMemBench.}
\label{tab:groupmembench_stats}
\end{table}

\subsubsection{Internal Enterprise-Chat Benchmark}
Our internal benchmark is constructed from anonymized enterprise group-chat
logs. It evaluates channel workflow, actor responsibility, artifact
lifecycle, cross-channel tracking, and multi-hop evidence retrieval.
Each question is annotated with an answer and supporting message evidence.

\begin{table}[!htbp]
\centering
\small
\resizebox{\columnwidth}{!}{
\begin{tabular}{llr}
\toprule
\textbf{Category} & \textbf{Statistic} & \textbf{Value} \\
\midrule
Corpus & Persistent channels & 3 \\
& Participants & 39 \\
& Messages & 10,222 \\
& Tracked artifact clusters & 22 \\
& Artifact-bearing messages & 1,158 \\
\midrule
Evaluation & Fine-grained Recall & 260 \\
& Memory Awareness & 263 \\
& Profile Understanding & 278 \\
& Total QA pairs & 801 \\
\bottomrule
\end{tabular}
}
\caption{Statistics and question distribution of the internal
enterprise-chat benchmark.}
\label{tab:internal_data}
\end{table}

\subsection{Implementation Details}
\label{sec:hyperparam}

This section details the implementation choices used for forest construction,
retrieval, and answer generation. Unless otherwise specified, we use
Qwen3-Embedding-0.6B as the frozen encoder for both message and node-summary
embeddings. Table~\ref{tab:implementation_details} summarizes the main
hyperparameters and dataset-specific answer-model settings.

\paragraph{Baseline Evaluation.}
We use Qwen3-Embedding-0.6B for EGCForest and for all reproduced baselines that require embeddings. On EverMemBench, we rerun BM25, Dense, BM25+Dense, GraphRAG, and A-Mem; results for MemoBase, Mem0, and MemOS are reported from the original benchmark paper. On GroupMemBench, we rerun all methods. All baselines on Company are evaluated by us. Because the filtered evaluation subsets used in prior studies were unavailable in the public releases we accessed, our reproduced runs use the full publicly available data for the corresponding task categories.

\paragraph{Forest Construction.}
All three forests use a common branching factor of $k=16$. Messages within
each tree are ordered chronologically and recursively grouped into
$k$-way internal nodes. This produces shallow trees while keeping the
input to each node-summary generation step manageable. For example, a tree
with approximately $10^4$ leaves has height four. The artifact construction
script has a default centroid-merge threshold of
$\theta_{\mathrm{merge}}=0.88$.

The following example shows an immutable log entry and its derived fields.
\begin{quote}
\small
\begin{tabular}{@{}ll@{}}
\texttt{msg\_id}     & $m_{091}$ \\
\texttt{timestamp}   & \texttt{2025-12-01 09:10} \\
\texttt{sender}      & \texttt{alice} \\
\texttt{channel}     & \texttt{\#backend} \\
\texttt{reply\_to}   & $m_{090}$ \\
\texttt{text}        & \parbox[t]{0.55\linewidth}{``\texttt{@Bob, PR for DOC-042 is merged.}''} \\
\texttt{mentions}    & \texttt{\{bob\}} \quad{\footnotesize (derived)} \\
\texttt{identifiers} & \texttt{\{DOC-042\}} \quad{\footnotesize (derived)} \\
\texttt{placements}  & \texttt{channel:\#backend} \\
                     & \texttt{actor:alice(auth)} \\
                     & \texttt{actor:bob(ment)} \\
                     & \texttt{artifact:DOC-042} \\
\end{tabular}
\end{quote}

\paragraph{Retrieval and Assembly.}
Table~\ref{tab:implementation_details} reports the default settings for
question answering, including a final context cap of 20 messages.
For the retrieval-budget sweep in Figure~\ref{fig:retrieval_pr}, we adjust
the final message cap to each evaluated cutoff $k \in \{20,30,40,50\}$.
The tree-level RRF constant $\eta_{\mathrm{tree}}=0$ gives reciprocal-rank
weights of $1/\mathrm{rank}$. The path and leaf limits in $S_2$
apply to candidate generation before final selection.

The expansion quota is shared by $S_2$ and $S_3$, including bridge messages.
Verbatim messages are included first;
summary headers use the remaining space, capped at 35\% of the total token
budget (1,050 tokens). This is an upper bound, not a reserved allocation.
Node summaries are precomputed and cached during construction, and online
LLM guidance is disabled; retrieval requires no LLM calls for query guidance,
memory traversal, or summarization.

\begin{table}[!htbp]
\centering
\footnotesize
\renewcommand{\arraystretch}{1.1}
\begin{tabular}{@{}>{\raggedright\arraybackslash}p{0.50\columnwidth}@{\hspace{0.04\columnwidth}}>{\raggedleft\arraybackslash}p{0.46\columnwidth}@{}}
\toprule
\textbf{Parameter} & \textbf{Value} \\
\midrule
Encoder & \texttt{Qwen3-Embedding-0.6B} \\
Branching factor $k$ & 16 \\
\midrule
RRF constants (message / tree) & 60 / 0 \\
Tree weights (Root / Leaf2tree / Collapsed) & 0.8 / 2.0 / 0.8 \\
$S_2$ paths / leaf cap per path & 8 / 4 \\
Bridge fanout / depth & 2 / 1 hop \\
Bridge messages/query (max.) & 2 \\
\midrule
$S_1$ / expansion quotas & 14 / 6 \\
Final messages (default max.) & 20 \\
Context budget & 3,000 tokens \\
Header budget (max.) & 35\% \\
\midrule
Answer (GroupMemBench) & GPT-5 \\
Answer (EverMemBench, Company) & GPT-4.1-mini \\
Judge (GroupMemBench) & GPT-5 \\
Judge (EverMemBench, Company) & Gemini-3-Flash-Preview \\
\bottomrule
\end{tabular}
\caption{Main implementation settings for \model{}.}
\label{tab:implementation_details}
\end{table}

\subsection{LLM Prompt Templates}
\label{sec:prompts}

We present condensed prompt templates. The answer-generation and grading
templates are condensed from the official \textsf{EverMemBench}
prompts~\citep{EverMemBench2026}. Multiple-choice questions require a single
uppercase option letter and are scored by direct matching; open-ended
answers use an LLM judge. Summaries are generated offline, and online
retrieval uses no LLM calls.

\paragraph{Node summaries.}
Inputs specify the view, tree label, interval, level, and ordered source
text: original messages at $L_1$, and consecutive child summaries above it.
\begin{quote}
\small
Write a compact index entry for routing to original evidence. Preserve
names, speaker attribution, dates, quantities, versions, statuses,
decisions, commitments, blockers, and unresolved questions. Keep events
chronological and express state changes as ``X $\rightarrow$ Y''.
Omit unsupported causes, outcomes, judgments, and advice; retain specific
names rather than replacing them with vague categories.

\emph{Channel:} identify participants and their contributions; keep
parallel workstreams distinct. For reply-thread inputs, preserve the
opening issue, reply context, and stated resolution.
\emph{Actor:} distinguish the person's own speech acts from others'
claims about them; preserve channel attribution and available evidence
IDs.
\emph{Artifact:} track lifecycle transitions, responsible speakers,
versions, statuses, decisions, blockers, and channel provenance.
\end{quote}
Non-root summaries use one paragraph, capped at 200/220/180 words for
Channel/Actor/Artifact; Actor summaries use 2--4 compact sentences.
Root cards use labelled prose: Channel records span, participants,
workstreams, decisions, and open threads; Actor records supported roles,
channels, responsibilities, explicit decision authority, observed phrasing,
others' attributed descriptions, and at most three evidence IDs; Artifact
records aliases, owners, channels, current status, lifecycle, and key
decisions. Unsupported fields are omitted. Actor root cards are capped
at 160 words.

\begin{samepage}
\paragraph{Open-ended answer generation.}
Input: retrieved memories and the question.
\begin{quote}
\small
Answer using only the supplied memories. Track who said what and when;
use direct evidence and prioritize the latest memory when accounts
conflict. Convert relative time expressions to specific dates using
message timestamps. Output only a concise answer, preferably under
5--6 words, without reasoning or extra explanation.
\end{quote}
\end{samepage}

\paragraph{Open-ended answer grading.}
Input: the question, gold answer, and generated answer.
\begin{quote}
\small
Judge whether the generated answer contains the gold answer's key
information. Accept equivalent wording and date formats; allow a
$\pm1$-day date difference for timezone variation. Return a JSON \texttt{label} of
\texttt{CORRECT} or \texttt{WRONG}.
\end{quote}

\subsection{Ethics Statement}
We hereby confirm that the entire work presented in this paper, including all content, analysis, and conclusions, is the original work of the human authors. While the core research, analysis, and conclusions are entirely the work of the human authors, generative AI was consulted for the limited purpose of improving the grammar and readability of the text.

\end{document}